\documentclass{SCIS2026}

\usepackage{array}
\usepackage[table]{xcolor}
\usepackage{makecell}
\usepackage{arydshln}
\usepackage{xspace}
\usepackage{overpic}
\usepackage{adjustbox}
\usepackage[section]{placeins}

\graphicspath{{figures/}}

\newcommand{\cmark}{\ding{51}}
\newcommand{\xmark}{\ding{55}}
\definecolor{green(pigment)}{rgb}{0.0, 0.65, 0.31}
\definecolor{red(pigment)}{rgb}{0.96, 0.59, 0.48}

\newcommand{\ua}{\mathbf{u}}
\newcommand{\sa}{\mathbf{s}}
\newcommand{\pa}{\mathbf{p}}
\newcommand{\ra}{\mathbf{r}}
\newcommand{\Lcal}{\mathcal{L}}
\newcommand{\Dcal}{\mathcal{D}}
\newcommand{\cowmlong}{Co-World Model}
\newcommand{\cowm}{CoWM\xspace}
\newcommand{\Zco}{\mathcal{Z}_{\rm co}}

\newsavebox{\pipeeqbox}
\newcommand{\pipeeq}[5]{%
  \put(#1,#2){\sbox{\pipeeqbox}{\resizebox*{#3\unitlength}{1.32\unitlength}{\fontfamily{ptm}\selectfont Eq.(\ref{#5})}}%
    \setlength{\fboxsep}{0.22\unitlength}%
    \hspace*{-0.22\unitlength}\colorbox[RGB]{#4}{\raisebox{\dp\pipeeqbox}{\usebox{\pipeeqbox}}}}}

\makeatletter
\g@addto@macro\normalsize{%
  \setlength\abovedisplayskip{7pt plus 2pt minus 3pt}%
  \setlength\belowdisplayskip{7pt plus 2pt minus 3pt}%
  \setlength\abovedisplayshortskip{0pt plus 2pt}%
  \setlength\belowdisplayshortskip{4pt plus 2pt minus 2pt}}
\makeatother

\begin{document}

\ArticleType{RESEARCH PAPER}
\Year{2026}
\Month{January}
\Vol{69}
\No{1}
\DOI{}
\ArtNo{}
\ReceiveDate{}
\ReviseDate{}
\AcceptDate{}
\OnlineDate{}
\AuthorMark{Li X T}
\AuthorCitation{Li X T, Jiang X, Liu C, et al}

\title{Co-World Model for Human-Video-to-Robot Transfer via Co-Constructed Shared Latent Dynamics}
      {Co-World Model for Human-Video-to-Robot Transfer via Co-Constructed Shared Latent Dynamics}

\author[1,2,4]{Xuetao LI}{}
\author[3]{Xuan JIANG}{{jiangxuan913@gmail.com}}
\author[4]{Chen LIU}{}
\author[2]{Zhongrui RAO}{}
\author[3]{\unskip\\Yu LING}{}
\author[3]{Hui WANG}{}
\author[5]{Sheng LIU}{}
\author[1,2]{Miao LI}{{miao.li@whu.edu.cn}}

\address[1]{School of Computer Science, Wuhan University, Wuhan 430072, China}
\address[2]{School of Robotics, Wuhan University, Wuhan 430072, China}
\address[3]{The 719th Research Institute of China State Shipbuilding Corporation Limited, Wuhan 430064, China}
\address[4]{Style3D Research, China}
\address[5]{Institute of Technological Sciences, Wuhan University, Wuhan 430072, China}

\abstract{Human videos are abundant and cheap to obtain, whereas robot trajectories remain scarce and costly, making zero-shot skill transfer from a single human demonstration an appealing route to scalable robot learning. Existing methods, however, either retarget human-specific geometry or match only static visual goals, and thus stumble over the cross-embodiment dynamics mismatch and the ungrounded action execution. We present the \emph{\cowmlong} (\cowm), a predictive framework jointly optimized from paired yet unsynchronized human and robot streams. To bridge the cross-embodiment dynamics gap, \cowm disentangles each observation into a shared object-relation state and embodiment-specific context. It aligns latent actions across embodiments using effect-swap and action-sensitivity objectives, such that actions are matched by their predicted effects on the manipulated objects rather than by embodiment-dependent kinematics. For grounded execution, a phase-aware receding-horizon controller estimates demonstration progress, projects the inferred human action prior onto the robot action-support density, and translates the resulting latent dynamics into executable robot commands. Extensive experiments in simulation and on two physical robot platforms demonstrate the effectiveness of CoWM, its cross-platform generalizability, and the contribution of each component. See the demo video in the supporting information for details.}

\keywords{world models, human video, robot learning, latent dynamics, action substitutability, cross-embodiment transfer, zero-shot imitation}

\maketitle


\section{Introduction}\label{sec:introduction}

\begin{figure}[!t]
\centering
\includegraphics[width=0.57\textwidth]{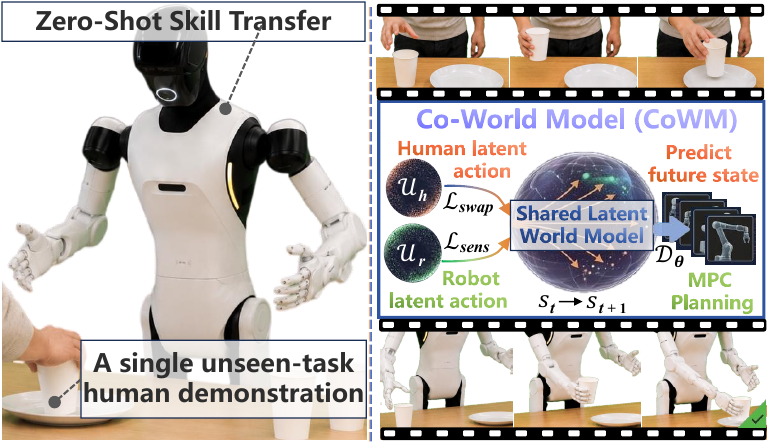}
\caption{\textbf{Human-video-to-robot zero-shot transfer with the \cowmlong.} Given a single unseen human demonstration (top film strip), the robot reasons in a co-constructed latent space where human and robot latent actions are bound to a shared latent world model by their predicted object effects, and grounds the induced plan into executable commands on its own embodiment (bottom film strip), acquiring the skill after watching the person once.}
\label{fig:teaser}
\end{figure}

Robot manipulation learning has advanced markedly, from imitation policies and vision-language-action models~\cite{diffusionpolicy,act,rt2} to generalist foundation policies that span tasks and platforms~\cite{pi0,geminirobotics,groot}.
These advances, however, are fueled by robot trajectories that remain scarce and costly, as teleoperated collection demands per-platform hardware, expert operators, and hours of supervised recording~\cite{openx,droid}.
Egocentric human videos offer the opposite profile, capturing abundant and easily acquired observations of contact-rich manipulation~\cite{egodex,dexcap}.
This asymmetry motivates \textit{human-video-to-robot zero-shot transfer}, in which a robot acquires an executable skill from a single unseen human demonstration without any corresponding robot trajectory, as illustrated in Figure~\ref{fig:teaser}.
However, most existing transfer methods either retarget human motion while preserving human-specific geometry~\cite{dexmv,handsasprobes,phantom}, or condition on visual goals or skill latents while discarding the intermediate interaction dynamics~\cite{track2act,vidman,uniskill}, and single-embodiment world models couple their predictive semantics to the one interface that generated the data~\cite{planet,dreamerv1,dreamerv3,tdmpc,tdmpc2}.

Realizing such transfer presents two unique challenges.
First, \textit{cross-embodiment dynamics mismatch}.
Humans and robots differ in morphology, sensing, and action interfaces, so different motions may produce the same object change while similar motions have state-dependent effects.
Latent proximity or marginal distribution matching cannot resolve this ambiguity, because a shared model may silently infer the embodiment, branch its transition rule, or ignore its action input.
Second, \textit{ungrounded action execution}.
Human videos carry neither robot commands nor proprioception, and executions proceed at mismatched rates.
A plan induced from the video alone may therefore drift out of phase or request actions the robot has never been observed to perform, and direct kinematic mapping cannot certify feasibility. In light of these challenges, this work investigates two critical questions:
\textbf{I) How can human observations and robot experience co-construct one predictive world model?}
\textbf{II) How can the latent plan induced by a single unseen human video be translated into feasible robot control at test time?}

In response to question \textbf{I)}, we propose the \emph{\cowmlong} (\cowm), jointly optimized from paired but unsynchronized human and robot streams.
\cowm factorizes each observation into a shared object-relation state and an embodiment-private context, and encodes both human and robot motion into a common latent action space.
Rather than aligning raw kinematics, we bind action semantics through predicted object effects with two complementary constraints, where an \textbf{effect-swap} loss requires human and robot actions with matched effects to predict compatible object changes, and an \textbf{action-sensitivity} margin rejects nearby actions whose observed effects differ.
Together they prevent the shared dynamics from branching on embodiment or degenerating into action-agnostic prediction.

Leveraging the co-constructed latent space, we design a \textbf{phase-aware grounded execution} scheme for question \textbf{II)}.
Online temporal matching tracks the demonstration phase against mismatched execution rates, a robot-state-conditioned residual corrects the human action prior, and an empirical action-support density confines proposals to actions supported by robot experience.
Receding-horizon planning under the shared dynamics then converts the demonstration-induced latent evolution into executable commands, replanning after each observation.

Our main contributions are summarized as follows:
\begin{itemize}
\item[\ding{182}] \textbf{\textit{Co-Constructed World Model.}} We formulate \cowm, in which human observations and robot experience jointly constrain a factorized predictive space over shared object relations and common latent actions, while embodiment-private variables retain the context tied to each embodiment's own body and controller.
\item[\ding{183}] \textbf{\textit{Effect-Grounded Action Semantics.}} We bind cross-embodiment latent actions through complementary effect-swap and action-sensitivity constraints, aligning actions by their predicted object effects rather than kinematics and rejecting action-agnostic shared dynamics.
\item[\ding{184}] \textbf{\textit{Grounded Zero-Shot Execution.}} We translate \cowm predictions into robot control through phase tracking, residual proposals, an action-support prior, and receding-horizon planning. Extensive experiments in simulation and on two physical robots validate the framework and the contribution of each component.
\end{itemize}

\section{Related Work}\label{sec:related}

\subsection{Robot Learning from Human Video}
Human video is a cheap and abundant substitute for costly robot trajectories, supplying visual rewards, affordances, goals, and reusable representations~\cite{r3mhuman,r3m,vip}. More direct routes retarget human motion, edit videos into robot-compatible training pairs, or exploit human play~\cite{dexmv,handsasprobes,phantom,mimicplay}, while point tracks, predicted motion, and egocentric hand annotations add structured supervision~\cite{track2act,vidman,egodex,dexcap}. These methods transfer geometry, task progress, or visual goals, yet none models how an action changes the task state, so the cross-embodiment dynamics mismatch is bypassed rather than resolved; \cowm instead lets both streams jointly constrain one predictive task world with an operational test of action substitutability.

\subsection{Demonstration-Conditioned Robot Policies}
Behavior cloning, diffusion policies, and action-chunking transformers learn expressive control from action-labeled robot demonstrations~\cite{bcz,diffusionpolicy,act}, and vision-language-action models scale this paradigm with language conditioning and large multi-task robot datasets~\cite{rt1,rt2,openvla,octo,pi0,openx,droid}. Related efforts target generalizable humanoid and dexterous manipulation~\cite{rgmp,tactilemllm,dexgrasp} and defend learned policies against backdoor attacks~\cite{attnbetray}. Their competence remains bounded by the same costly robot data that motivates our setting, and they are not conditioned on a held-out human video or execution phase. Our evaluation therefore includes demonstration-conditioned variants that receive the same human observations, paired training split, and robot state as \cowm, isolating the shared predictive world rather than the information budget.

\begin{figure}[!t]
\centering
\begin{overpic}[width=\textwidth]{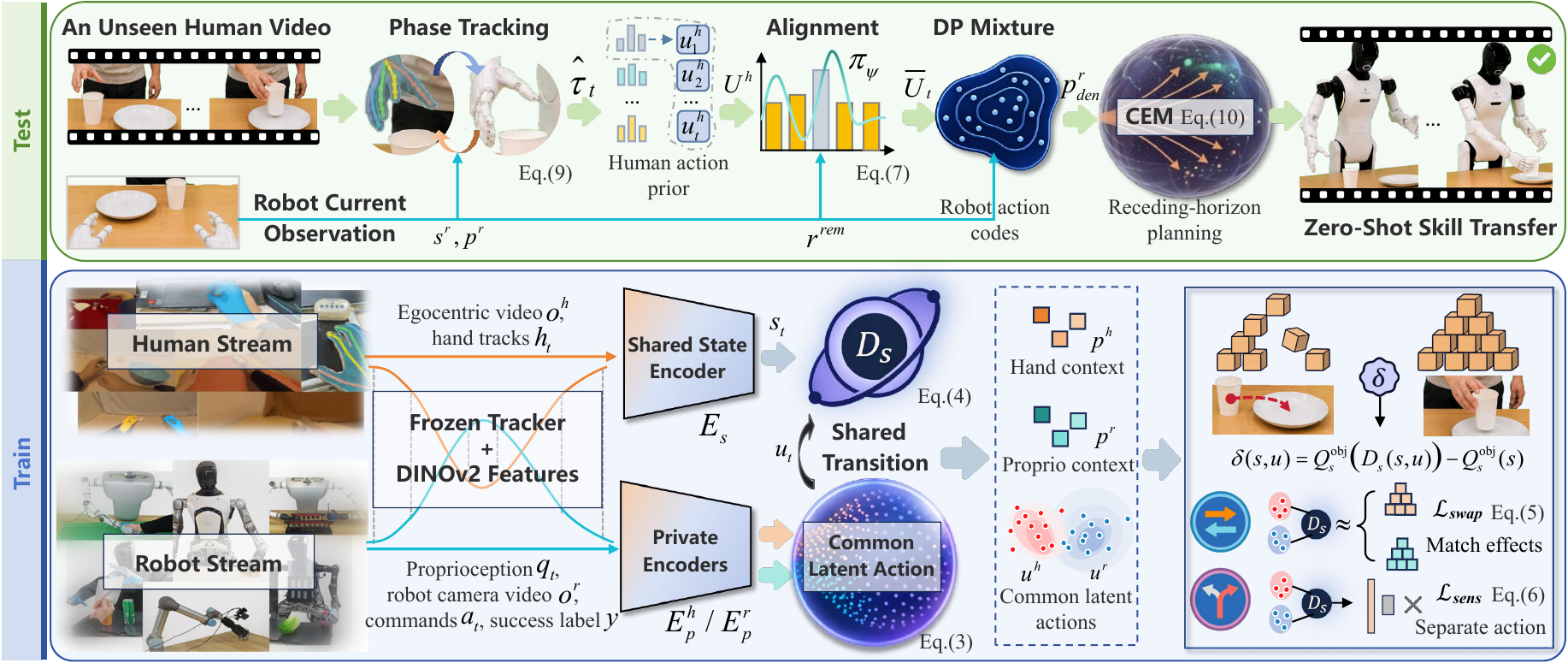}
    \pipeeq{75.31}{33.93}{4.13}{205,203,209}{eq:mpc}
    \pipeeq{33.08}{30.34}{3.44}{244,250,244}{eq:phase}
    \pipeeq{54.61}{30.34}{3.44}{244,250,244}{eq:grounding}
    \pipeeq{58.53}{16.14}{3.44}{241,245,252}{eq:dynamics}
    \pipeeq{58.64}{0.50}{3.45}{241,245,252}{eq:action}
    \pipeeq{95.10}{8.70}{3.44}{241,245,252}{eq:swap}
    \pipeeq{95.15}{3.45}{3.45}{241,245,252}{eq:sensitivity}
\end{overpic}
\caption{\textbf{Overview of the \cowmlong.} \emph{Training} (bottom): a frozen tracker with DINOv2 features factorizes paired yet unsynchronized human and robot streams into a shared object-relation state $\sa_t$ with embodiment-private contexts $\pa^h,\pa^r$ (Eq.~\eqref{eq:state}), while motion encoders map both streams into common latent actions $\ua^h,\ua^r$ (Eq.~\eqref{eq:action}). The shared transition $D_s$ (Eq.~\eqref{eq:dynamics}) yields predicted object effects $\delta(\sa,\ua)$ (Eq.~\eqref{eq:objecteffect}), which the effect-swap loss (Eq.~\eqref{eq:swap}) matches across embodiments while the action-sensitivity margin (Eq.~\eqref{eq:sensitivity}) separates different-effect actions. \emph{Test} (top): given a single unseen human video, phase tracking (Eq.~\eqref{eq:phase}) selects the human action prior $U^h$, a residual proposal corrects it with the robot state (Eq.~\eqref{eq:grounding}), and receding-horizon CEM planning (Eq.~\eqref{eq:mpc}) under the DP action-support density $p_{\rm den}^r$ converts the shared latent evolution into executable commands, yielding zero-shot skill transfer.}
\label{fig:pipeline}
\end{figure}

\subsection{Cross-Embodiment Latent World Models}
Latent world models plan through learned state transitions~\cite{planet,dreamerv1,dreamerv3,tdmpc,tdmpc2}, predictive video models serve as planners or policies~\cite{unipi,gr1,genie}, and latent-action methods infer control variables from observation-only sequences~\cite{ilpo,vpt,lapo}. Recent cross-embodiment systems condition robot policies on skill representations from human video~\cite{uniskill} or contrastively align end-effector actions through kinematic retargeting~\cite{lad}, yet their alignment still rests on frame prediction or kinematic correspondence, not on what an action does to the task state. \cowm instead co-constructs an object-centric predictive state and a common latent action interface from both streams; since a shared encoder or transition alone may retain domain cues, branch on embodiment, or ignore its action input, the effect-swap and action-sensitivity constraints directly target these failure modes. On the execution side, SoftDTW~\cite{softdtw}, MMD~\cite{mmd}, Dirichlet-process mixtures~\cite{ferguson1973,bleijordan2006}, and CEM~\cite{cem} are established components assembled to ground the plan; the effect-constrained shared dynamics remain the central mechanism.

\section{Co-World Model}\label{sec:method}

This section instantiates the two responses outlined in the introduction, following the pipeline in Figure~\ref{fig:pipeline}. To resolve the cross-embodiment dynamics mismatch, we first formalize the co-construction objective, then factorize observations into shared and private states and bind latent actions through effect-grounded dynamics (Figure~\ref{fig:pipeline}, bottom). To resolve the ungrounded action execution, we subsequently attach robot grounding with an action-support prior and perform phase-aware receding-horizon planning in the co-constructed latent space (Figure~\ref{fig:pipeline}, top). The two stages communicate only through the co-constructed latent space and the grounded decoder, so the test-time planner consumes exactly the quantities that training was optimized to provide.

\subsection{Human--Robot Co-Construction Objective}\label{sec:objective}
Training uses paired but unsynchronized segments $\Dcal=\{(\tau_n^h,\tau_n^r,y_n)\}_{n=1}^N$. The embodiment index is $e\in\{h,r\}$, and $y_n\in\{0,1\}$ records if the scripted replay succeeds. The human stream contains egocentric RGB $o_t^h$ and hand tracks $h_t$ (tracker illustrated in Figure~\ref{fig:handtrack}); the robot stream contains RGB $o_t^r$, proprioception $q_t$, normalized $7$D end-effector command $a_t$, and $y_n$. We write $\mathrm{sg}(\cdot)$ for stop-gradient and $\mathbb I[\cdot]$ for the indicator. At test time, a held-out human video has no paired robot trajectory, teleoperation, or scripted-controller call.

The factorized \cowm comprises a shared object-centric task state $\sa$, a common latent action $\ua$, shared task dynamics $D_s$, and embodiment-private context $\pa^e$ for cross-embodiment prediction. We denote the shared predictive state--action space by $\Zco=\mathcal S_{\rm obj}\times\mathcal U$. The task state $\sa\in\mathcal S_{\rm obj}$ contains shared object-relation content, and the shared latent-action coordinates $\ua\in\mathcal U$ are populated by embodiment-specific motion encoders. The same transition $D_s:\Zco\rightarrow\mathcal S_{\rm obj}$ predicts object-state evolution for both embodiments. Human and robot streams jointly constrain its coordinates and transition law, while the embodiment-private variables $\pa^e$ retain sensing, morphology, reconstruction, and control context that should not be aligned and are not part of $\Zco$. In this paper, \emph{co-construction} denotes joint offline optimization of the shared $\sa$, $\ua$, and $D_s$ structure from paired but unsynchronized human and robot streams; it does not denote synchronized control, online human--robot interaction, or identical observation and action interfaces.

For a learned task state $\sa$, define the predicted object effect
\begin{equation}
\delta(\sa,\ua)=Q_s^{\rm obj}(D_s(\sa,\ua))-Q_s^{\rm obj}(\sa).
\label{eq:objecteffect}
\end{equation}
Actions $\ua^h$ and $\ua^r$ are locally substitutable at $\sa$ when their predicted effects under shared dynamics differ by at most a task tolerance. This is the operational consistency condition used to bind human and robot action semantics inside $\Zco$. The criterion must also separate a robot action whose different effect was observed at a nearby object state. It is local and model-dependent over sampled states and executable actions; the negative relies on local effect-label stability. It does not identify a unique latent coordinate system or imply a physical counterfactual at a human state.

\subsection{Shared World State and Private Context}\label{sec:state}
Co-construction first requires comparable task state without erasing embodiment-specific control information. One frozen tracker supplies role-ordered masks $m_{t,k}^e$ and confidences $\chi_{t,k}^e$ in both domains. Paired annotations may replace tracker outputs only when both streams have the same provenance; primary test inference always uses the tracker. Let $F_t^e$ be a DINOv2 ViT feature map~\cite{dinov2,vit,gsa}, and let $\ra_t^e=[\ra_{t,\rm obj}^e,\ra_{t,\rm exec}^e]$ collect normalized object and effector relations. We factor each observation into shared task state and embodiment-private state:
\begin{equation}
\sa_t^e=E_s(\{\mathrm{Pool}(F_t^e,m_{t,k}^e)\}_{k\in\rm obj},\ra_{t,\rm obj}^e),\quad
\pa_t^h=E_p^h(F_t^h,h_t,\ra_{t,\rm exec}^h),\quad
\pa_t^r=[q_t,E_p^r(F_t^r,\ra_{t,\rm exec}^r)].
\label{eq:state}
\end{equation}
Thus only the object-centric task state $\sa$ is architecturally shared. The private variable $\pa^e$ preserves information needed for reconstruction and feasible robot control; \cowm therefore does not assume that the entire latent representation is embodiment-invariant. The heads $Q_s^{\rm obj}$ and $Q_q$ recover object relations and robot proprioception, respectively.

\begin{figure}[!t]
\centering
\includegraphics[width=0.58\textwidth]{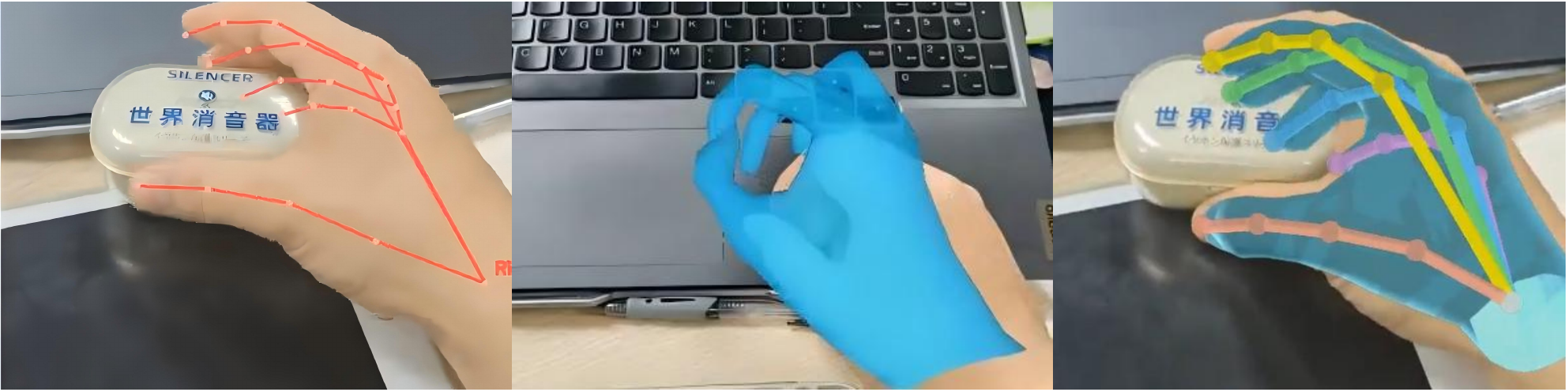}
\caption{\textbf{Hand tracking for the human stream.} Left: joint-angle-only tracking adopted by prior methods. Middle: MANO-based hand mesh estimation. Right: our tracker combines the joint-angle estimates with the MANO mesh to produce the hand tracks $h_t$ consumed by \cowm{}.}
\label{fig:handtrack}
\end{figure}

Define local pose increments $\Delta T_t^x=\log((T_t^x)^{-1}T_{t+1}^x)$ and pinch aperture $c_t=\|p_t^{\rm index}-p_t^{\rm thumb}\|_2$. Embodiment-specific encoders then map human and robot motion histories into the common CoWM action interface:
\begin{equation}
\mathbf d_t^h=[\Delta T_t^h,c_t],\quad
\mathbf d_t^r=[\Delta T_t^{ee},g_t],\quad
\ua_t^e=g_e(\mathbf d_{t-L:t}^e).
\label{eq:action}
\end{equation}
Here $T^h$ and $T^{ee}$ are wrist and end-effector poses, and $g_t$ is gripper state. Every channel is scaled with training-split statistics; missing history is edge-padded. Hand poses come from annotations or a fixed estimator~\cite{mediapipe}. Figure~\ref{fig:handtrack} illustrates how the hand tracks $h_t$ are obtained. Prior pipelines rely on joint angles alone (left) or on a MANO mesh alone (middle); our tracker fuses the two estimates (right), so the motion descriptors $\mathbf d_t^h$ inherit both the articulation detail of the joint angles and the surface geometry of the mesh.

Both streams jointly optimize the CoWM representation objective $\Lcal_{\rm repr}$. State reconstruction constrains $(\sa,\pa)$ while masks are decoded from $\sa$ alone; inverse dynamics anchors $\ua_t^e$ to observed transitions; motion reconstruction and a variance floor rule out the constant-code solution. The task-state loss applies every term to both embodiments:
\begin{equation}
\Lcal_{\rm state}=\sum_{e,t}\Big[
\|o_t^e-G_o^e(\sa_t^e,\pa_t^e)\|_1
+\lambda_{\rm rel}\|Q_s(\sa_t^e)-\ra_{t,\rm obj}^e\|_2^2
+\lambda_{\rm mask}\operatorname{BCE}
(B_s(\sa_t^e),m_{t,\rm obj}^e)\Big].
\label{eq:state-loss}
\end{equation}
$B_s$ reads only the task-state branch. An entity-validity mask removes padded or missing objects from the relation and mask terms.

The observed-transition inverse anchor is
\begin{equation}
\Lcal_{\rm inv}=\sum_{e,t}
\|\ua_t^e-\operatorname{IDM}_\eta(\sa_t^e,\sa_{t+1}^e)\|_2^2.
\end{equation}
Motion reconstruction is
\begin{equation}
\Lcal_{\rm mot}=\sum_{e,t}\|R_e(\ua_t^e)-\mathbf d_t^e\|_2^2.
\end{equation}
For action coordinate $c$, define $\sigma_c^e=\operatorname{Std}_{\mathcal B}([\ua^e]_{:,c})$ and $\sigma_c^{hr}=\operatorname{Std}_{\mathcal B}([\ua^h;\ua^r]_{:,c})$. The variance floor is
\begin{equation}
\Lcal_{\rm var}=\sum_{e,c}[\nu-\sigma_c^e]_+^2
+\sum_c[\nu-\sigma_c^{hr}]_+^2.
\end{equation}
The combined \cowm{} representation objective is
\begin{equation}
\Lcal_{\rm repr}=\Lcal_{\rm state}+\lambda_{\rm act}
(\Lcal_{\rm inv}+\lambda_{\rm mot}\Lcal_{\rm mot}
+\lambda_{\rm var}\Lcal_{\rm var}).
\end{equation}

Paired executions need not be synchronized. SoftDTW aligns relation progress and detached action distance into a confidence-weighted monotone path $\widetilde\Gamma$~\cite{softdtw}. Action distance is disabled during representation warm-up. The resulting pair loss $\Lcal_{\rm seed}$ only initializes CoWM coordinates, and its weight is annealed to zero before effect training; otherwise exact code equality would make substitutability trivial. The final objective retains only MK-MMD as a weak distribution regularizer~\cite{mmd}. Neither term establishes shared predictive semantics. 

For pair $n$, set $\bar\ra_{n,t}^e=\ra_{n,t}^e-\ra_{n,1}^e$ and
\begin{equation}
C_{n,ij}=\beta_r\|\bar\ra_{n,i}^h-\bar\ra_{n,j}^r\|_2^2
+\beta_u\|\operatorname{sg}(\ua_{n,i}^h-\ua_{n,j}^r)\|_2^2.
\end{equation}
Let $A_n=\operatorname{sg}(\nabla_C\operatorname{SoftDTW}_\gamma(C_n))$ and $\Gamma_{n,ij}=A_{n,ij}/(\sum_{p,q}A_{n,pq}+\epsilon)$. With $c_{n,i}^e=\min_{k\in\mathrm{obj}}\chi_{n,i,k}^e$, the confidence-weighted path is
\begin{equation}
\widetilde\Gamma_{n,ij}=\frac{\Gamma_{n,ij}c_{n,i}^hc_{n,j}^r}
{\sum_{p,q}\Gamma_{n,pq}c_{n,p}^hc_{n,q}^r+\epsilon}.
\end{equation}
Pairs with zero confidence mass are excluded and counted. The finite warm-up seed is
\begin{equation}
\Lcal_{\rm seed}=\sum_{n:y_n=1}\sum_{i,j}\widetilde\Gamma_{n,ij}
\|\ua_{n,i}^h-\ua_{n,j}^r\|_2^2.
\end{equation}
MK-MMD uses the unbiased U-statistic with validation-fixed kernel
\begin{equation}
k(x,y)=\sum_m\beta_m
\exp\!\left(-\frac{\|x-y\|_2^2}{2\sigma_m^2}\right).
\end{equation}
The fixed weights satisfy $\beta_m\geq0$ and $\sum_m\beta_m=1$.

\subsection{Effect-Grounded Co-World Dynamics}\label{sec:dynamics}
The shared transition $D_s$ is the predictive core of \cowm (Figure~\ref{fig:pipeline}, bottom right): both embodiments train it to evolve object-centric task state under latent actions. Robot-private dynamics $D_p^r$ separately preserve configuration needed for grounding and feasibility:
\begin{equation}
\tilde\sa_{t,\ell+1}^e=D_s(\tilde\sa_{t,\ell}^e,\ua_{t+\ell}^e),\qquad
\tilde\pa_{t,\ell+1}^r=D_p^r(\tilde\pa_{t,\ell}^r,\tilde\sa_{t,\ell}^r,\ua_{t+\ell}^r).
\label{eq:dynamics}
\end{equation}
The rollout starts from the encoded observation. A discounted $H_{\rm train}$-step loss $\Lcal_{\rm dyn}^{H_{\rm train}}$ penalizes shared-state prediction in both domains and private-state prediction for the robot. The first term learns object evolution; the second preserves configuration for decoding and feasibility. For valid starts, let $\epsilon_{n,t,\ell}^{s,e}=\tilde\sa_{n,t,\ell}^e-\operatorname{sg}(\sa_{n,t+\ell}^e)$ and $\epsilon_{n,t,\ell}^{p}=\tilde\pa_{n,t,\ell}^r-\operatorname{sg}(\pa_{n,t+\ell}^r)$. The multi-step loss is
\begin{equation}
\Lcal_{\rm dyn}^{H_{\rm train}}=
\sum_{n,e,t}\sum_{\ell=1}^{H_{\rm train}}
\gamma_{\rm roll}^{\ell-1}\|\epsilon_{n,t,\ell}^{s,e}\|_2^2
+\lambda_p\sum_{n,t}\sum_{\ell=1}^{H_{\rm train}}
\gamma_{\rm roll}^{\ell-1}\|\epsilon_{n,t,\ell}^{p}\|_2^2.
\end{equation}

Training one transition function on both streams is not sufficient to resolve the dynamics mismatch, since the model may still branch on embodiment or ignore its action input. For a successful pair, set $t_h=i$, $t_r=j$, $w_{n,ij}=\widetilde\Gamma_{n,ij}$, and $\delta_{n,t}^e(\ua)=\delta(\sa_{n,t}^e,\ua)$. The swap loss therefore evaluates both codes from each aligned start state:
\begin{equation}
\Lcal_{\rm swap}=\sum_{n,i,j}w_{n,ij}\sum_{e\in\{h,r\}}
\|\delta_{n,t_e}^e(\ua_{n,i}^h)-\delta_{n,t_e}^e(\ua_{n,j}^r)\|_2^2.
\label{eq:swap}
\end{equation}
This is a predictive constraint; physical evidence still requires identical-state robot interventions.

A swap loss alone admits an action-independent model. We therefore sample a detached robot action from the same task and a nearby object state, but with a different observed effect. Let $d_+^e$ and $d_-^e$ denote next-object prediction errors for the aligned and negative codes. The margin
\begin{equation}
\Lcal_{\rm sens}=\sum_{(n,i,j)\in\mathcal I^-}w_{n,ij}\sum_e
[\mu_{\rm sens}+d_+^e-d_-^e]_+.
\label{eq:sensitivity}
\end{equation}
is positive if the model ignores action. $\mathcal I^-$ contains the aligned transitions of successful pairs for which both domain-specific negative sets are nonempty and both successors are observed; empty sets are skipped. For transition $(n,e,t)$, nearby-state negatives are drawn from
\begin{equation}
\mathcal B_{n,e,t}^{-}=\{b:\;y_{n(b)}=1,\ \mathrm{task}(b)=\mathrm{task}(n),\ \|\ra_{b,\rm obj}^r-\ra_{n,t,\rm obj}^e\|_2\leq\epsilon_s,\ \|\Delta\ra_{b,\rm obj}^r-\Delta\ra_{n,t,\rm obj}^e\|_2
\geq\epsilon_\delta\}.
\end{equation}
One code is sampled by a seed-fixed rule and detached; empty sets are skipped. This construction assumes effect labels are locally stable, which the equal-effect intervention control must test.

For an aligned pair $(n,i,j)$, set $t_h=i$ and $t_r=j$, with positive codes $\ua_{n,ij,+}^h=\ua_{n,i}^h$ and $\ua_{n,ij,+}^r=\ua_{n,j}^r$. For each domain $e$, let $b_e$ be sampled from $\mathcal B_{n,e,t_e}^{-}$ and set $\ua_{n,ij,-}^e=\operatorname{sg}(\ua_{b_e}^r)$. The positive and negative successor errors are
\begin{equation}
d_{n,ij,\pm}^e=
\|Q_s^{\rm obj}(D_s(\sa_{n,t_e}^e,\ua_{n,ij,\pm}^e))
-\ra_{n,t_e+1,\rm obj}^e\|_2^2.
\end{equation} The inverse-forward anchor $\Lcal_{\rm if}$ and observed-delta anchor $\Lcal_{\rm trans}$ provide additional grounding:
\begin{equation}
\Lcal_{\rm if}=\sum_{e,t}\|\operatorname{IDM}_\eta
(\sa_t^e,D_s(\sa_t^e,\ua_t^e))-\operatorname{sg}(\ua_t^e)\|_2^2.
\end{equation}
With $\Delta Q_t^e=Q_s^{\rm obj}(\sa_{t+1}^e)-Q_s^{\rm obj}(\sa_t^e)$, the observed-effect anchor is
\begin{equation}
\Lcal_{\rm trans}=\sum_{n:y_n=1}\sum_{i,j}
\widetilde\Gamma_{n,ij}\|\Delta Q_{n,i}^h-\Delta Q_{n,j}^r\|_2^2.
\end{equation}
The grouped effect objective used in the training schedule is
\begin{equation}
\Lcal_{\rm effect}=\Lcal_{\rm swap}+\lambda_{\rm sens}\Lcal_{\rm sens}
+\lambda_{\rm if}\Lcal_{\rm if}+\lambda_{\rm trans}\Lcal_{\rm trans}.
\end{equation}

Together, swap and sensitivity provide CoWM's cross-embodiment semantic binding on sampled local support: the first equates matched predicted effects, and the second rejects different-effect actions. They neither identify a unique latent coordinate system nor guarantee transfer outside observed states. The prespecified $2\times2$ intervention is designed to test their non-redundant roles.

\noindent\textbf{Why substitution and sensitivity are complementary.} The role of the two constraints can be stated on finite training support. The dynamics loss evaluates observed inputs $(\sa_i^h,\ua_i^h)$ and $(\sa_j^r,\ua_j^r)$, but not the crossed inputs $(\sa_i^h,\ua_j^r)$ and $(\sa_j^r,\ua_i^h)$. For any dynamics model that fits the observed inputs, another model can agree on every observed input and assign arbitrary effects to the crossed inputs; it has the same $\Lcal_{\rm dyn}$ and $\Lcal_{\rm MMD}$. Action-code proximity alone does not exclude this construction without an independently enforced continuity bound. $\Lcal_{\rm swap}$ directly constrains these crossed predictions on sampled pairs. The pairwise $\Lcal_{\rm seed}$ is therefore used only for initialization and reaches zero weight before these constraints are optimized; retaining it to exact equality would make substitution redundant.

Substitution alone is also insufficient. If $D_s(\sa,\ua)=f(\sa)$ ignores action, the two swapped effects are identical and $\Lcal_{\rm swap}=0$. For a valid different-effect negative, however, $d_+=d_-$ and its sensitivity term equals the positive margin $\mu_{\rm sens}$. Thus substitution removes unconstrained crossed completions, while sensitivity rejects the action-independent completion. This argument is local to the sampled support and does not establish global substitutability or latent identifiability.

\subsection{Robot Grounding and Action-Support Prior}\label{sec:grounding}
Resolving the ungrounded action execution first requires knowing which latent actions the robot has actually been observed to perform. To this end, a truncated stick-breaking DP mixture~\cite{ferguson1973,bleijordan2006} is fitted to successful detached action codes; robot responsibilities produce the planning density $p_{\rm den}^r$. This auxiliary density is a global empirical density, not success probability, strict support, or state-conditioned feasibility, and occupied components are not interpreted as skills. For $k<K_{\max}$, $v_k\sim\operatorname{Beta}(1,\alpha)$, while $v_{K_{\max}}=1$ and $\omega_k=v_k\prod_{k'<k}(1-v_{k'})$. For pooled successful action $b$,
\begin{equation}
c_b\sim\operatorname{Cat}(\boldsymbol\omega),\qquad
\ua_b\sim\mathcal N(\mu_{c_b},\Sigma_{c_b}).
\end{equation}
Each coordinate uses a Normal--Inverse-Gamma base measure $G_0(m_0,\kappa_0,a_0,b_0)$. The mean-field family is
\begin{equation}
q(v,c,\mu,\Sigma)=\prod_{k<K_{\max}}q(v_k)
\prod_b q(c_b)\prod_k q(\mu_k,\Sigma_k),
\end{equation}
where $q(v_k)$ is Beta, $q(c_b)=\operatorname{Cat}(\boldsymbol\zeta_b)$, and component factors are Normal--Inverse-Gamma. Let $U=\{\operatorname{sg}(\ua_b)\}_b$ be the detached successful action set. We minimize
\begin{equation}
\Lcal_{\rm DP}=-\mathbb E_q[\log p(U,c,v,\mu,\Sigma)
-\log q(c,v,\mu,\Sigma)].
\end{equation}
Posterior means of the component locations and diagonal covariances are used in the plug-in density. Robot occupancy weights are
\begin{equation}
\bar\omega_k^r=\frac{\sum_{b:e_b=r}\zeta_{bk}}
{\sum_{k'}\sum_{b:e_b=r}\zeta_{bk'}+\epsilon}.
\end{equation}
The resulting robot planning density is
\begin{equation}
p_{\rm den}^r(\ua)=\sum_k\bar\omega_k^r
\mathcal N(\ua\mid\mu_k,\Sigma_k).
\end{equation}

SoftDTW-weighted human chunks $\breve U^h$ and remaining object progress $\ra^{\rm rem}$ condition a robot-state-aware correction:
\begin{equation}
\tilde U^\pi=\mathrm{sg}(\breve U^h)+
\pi_\psi(\sa^r,\pa^r,\ra^{\rm rem},\mathrm{sg}(\breve U^h),M).
\label{eq:grounding}
\end{equation}
Human actions and object relations aligned to robot index $j$ are
\begin{equation}
\breve\ua_{n,j}^h=\frac{\sum_i\widetilde\Gamma_{n,ij}\ua_{n,i}^h}
{\sum_i\widetilde\Gamma_{n,ij}+\epsilon},\qquad
\breve\ra_{n,j,\rm obj}^h=\frac{\sum_i\widetilde\Gamma_{n,ij}\ra_{n,i,\rm obj}^h}
{\sum_i\widetilde\Gamma_{n,ij}+\epsilon}.
\end{equation}
A fixed $H_{\rm MPC}$ chunk is zero-padded with mask $M_{n,j,\ell}=\mathbb I[j+\ell<T_n^r]$, and $\ra^{\rm rem}=\ra_{T_h,\rm obj}^h-\breve\ra_{n,j,\rm obj}^h$. Here $T_n^r$ counts observation frames, so transition actions end at index $T_n^r-1$.
The decoder maps a selected code to the robot command, $\hat a_t=A_r(\ua_t,\sa_t^r,\pa_t^r)$. Success-weighted losses $\Lcal_{\rm dec}^w$ and $\Lcal_{\rm prop}^w$ regress recorded commands and aligned robot latent chunks. Let $w_n=y_n+\rho_f(1-y_n)$. The decoder loss is
\begin{equation}
\Lcal_{\rm dec}^w=\sum_{n,t}w_n
\|A_r(\operatorname{sg}(\ua_{n,t}^r),\sa_{n,t}^r,\pa_{n,t}^r)-a_{n,t}\|_2^2.
\end{equation}
The residual-proposal loss is
\begin{equation}
\Lcal_{\rm prop}^w=\sum_{n,j,\ell}w_nM_{n,j,\ell}
\|\tilde\ua_{n,j,\ell}^{\pi}-\operatorname{sg}(\ua_{n,j+\ell}^r)\|_2^2.
\end{equation}
The grouped grounding objective is
\begin{equation}
\Lcal_{\rm grd}=\Lcal_{\rm dec}^w+\lambda_\pi\Lcal_{\rm prop}^w.
\end{equation} Stop-gradients prevent grounding from redefining the action encoders.

We group swap, sensitivity, and the two transition anchors as $\Lcal_{\rm effect}$, and decoder/proposal supervision as $\Lcal_{\rm grd}$. The complete objective jointly trains the factorized CoWM representation, shared dynamics, cross-embodiment semantic constraints, and robot grounding:
\begin{equation}
\Lcal_{\rm neural}=\Lcal_{\rm repr}
+\lambda_{\rm dyn}\Lcal_{\rm dyn}^{H_{\rm train}}
+\lambda_{\rm MMD}\Lcal_{\rm MMD}
+\lambda_{\rm eff}\Lcal_{\rm effect}+\lambda_{\rm grd}\Lcal_{\rm grd}.
\label{eq:total}
\end{equation}
DP factors each minimize $\Lcal_{\rm DP}$ on detached codes and therefore cannot redefine the representation. Training proceeds through representation/dynamics warm-up, effect training after the weight on $\Lcal_{\rm seed}$ reaches zero, and joint fine-tuning.

\subsection{Planning in the Co-World Latent Space}\label{sec:planning}
Grounded execution finally requires staying synchronized with a demonstration that proceeds at its own rate (Figure~\ref{fig:pipeline}, top). At robot time $t$, CoWM planning therefore first estimates demonstration phase from object-state history only. Define $\boldsymbol\varphi_t^e=[\ra_{t,\rm obj}^e-\ra_{1,\rm obj}^e,\,\ra_{t,\rm obj}^e-\ra_{t-1,\rm obj}^e]$, with the second term zero at the first frame, and let $\Phi_{1:t}^e$ be its history. The monotone estimate is
\begin{equation}
\hat\tau_t=\arg\min_{\tau\in\mathcal W_t}
d_{\rm ph}(\Phi_{1:t}^r,\Phi_{1:\tau}^h),
\label{eq:phase}
\end{equation}
where $\mathcal W_t=[\hat\tau_{t-1},\min(T_h,\hat\tau_{t-1}+W)]$ and $\hat\tau_0=1$. The path-weighted average SoftDTW cost $d_{\rm ph}$ makes prefix lengths comparable while the search window prevents phase reversal. For phase cost $C_{pq}^{\rm ph}=\|\boldsymbol\varphi_p^r-\boldsymbol\varphi_q^h\|_2^2$ and SoftDTW path derivative $P^{\rm ph}$,
\begin{equation}
d_{\rm ph}=\frac{\sum_{p,q}P_{pq}^{\rm ph}C_{pq}^{\rm ph}}
{\sum_{p,q}P_{pq}^{\rm ph}+\epsilon}.
\end{equation} The phase selects the human prior $U_t^h=[\ua_{\hat\tau_t+\ell}^h]_{\ell=0}^{H_t-1}$, where $H_t=\min(H_{\rm MPC},T_h-\hat\tau_t)$. The residual proposal (Eq.~\eqref{eq:grounding}) produces $\tilde U_t^\pi$ from $U_t^h$, the current robot state, and remaining object progress. CEM~\cite{cem} samples near $\tilde U_t^\pi$ under a DP mixture reweighted by human component responsibility and robot occupancy. This keeps proposals human-conditioned while avoiding modes unsupported by robot data. For human action $\ua_{\hat\tau_t+\ell}^h$, define its mixture responsibility
\begin{equation}
\zeta_{t,\ell,k}^h\propto\mathbb E_q[\omega_k]
\mathcal N(\ua_{\hat\tau_t+\ell}^h\mid\mu_k,\Sigma_k).
\end{equation}
Combining this responsibility with robot occupancy gives
\begin{equation}
\tilde\omega_{t,\ell,k}^r=
\frac{\zeta_{t,\ell,k}^h\bar\omega_k^r}
{\sum_{k'}\zeta_{t,\ell,k'}^h\bar\omega_{k'}^r+\epsilon},\qquad
p_{\rm mix,\ell}^r(\ua)=\sum_k\tilde\omega_{t,\ell,k}^r
\mathcal N(\ua\mid\mu_k,\Sigma_k).
\end{equation}
The initial CEM density is proportional to
$\mathcal N(\ua\mid\tilde\ua_{t,\ell}^{\pi},\Sigma_0)p_{\rm mix,\ell}^r(\ua)$, where $\tilde\ua_{t,\ell}^{\pi}$ is the $\ell$-th entry of the test-time residual-corrected proposal $\tilde U_t^\pi$.

Starting from $(\hat\sa_0,\hat\pa_0)=(\sa_t^r,\pa_t^r)$, the CoWM dynamics in Eq.~\eqref{eq:dynamics} predict each candidate in the factorized shared/private state. The task cost $J_{\rm task}$ combines human-trajectory tracking and terminal-goal error; $J_{\rm reg}$ combines deviation from $\tilde U_t^\pi$, DP density, and command smoothness. With $\ra_\ell^{\rm ref}=\ra_{t,\rm obj}^r+\ra_{\hat\tau_t+\ell,\rm obj}^h-\ra_{\hat\tau_t,\rm obj}^h$ and $\ra_g=\ra_{t,\rm obj}^r+\ra_{T_h,\rm obj}^h-\ra_{\hat\tau_t,\rm obj}^h$, the MPC costs are
\begin{gather}
J_{\rm traj}=\sum_{\ell=1}^{H_t}
\|Q_s^{\rm obj}(\hat\sa_\ell)-\ra_\ell^{\rm ref}\|_2^2,\qquad
J_{\rm prior}=\sum_{\ell=0}^{H_t-1}
\|\ua_\ell-\tilde\ua_{t,\ell}^{\pi}\|_2^2,\qquad
J_{\rm den}=-\sum_{\ell=0}^{H_t-1}\log p_{\rm den}^r(\ua_\ell),\\*
J_{\rm sm}=\sum_{\ell=0}^{H_t-1}\|\hat a_\ell-\hat a_{\ell-1}\|_2^2,\qquad
J_{\rm goal}=\|Q_s^{\rm obj}(\hat\sa_{H_t})-\ra_g\|_2^2.
\end{gather}
The previous command $\hat a_{-1}$ is the last executed command, or a hold command at $t=1$. CEM solves
\begin{equation}
U_t^*=\arg\min_U\,[J_{\rm task}(U)+J_{\rm reg}(U)].
\label{eq:mpc}
\end{equation}
CEM uses validation-frozen samples, elite count, iterations, covariance floor, horizon, and bounds. If fewer than the required elite count are feasible, it refits to feasible samples only; with no feasible sample it executes hold and records a planning failure. When $H_t=0$, the terminal predicate is checked without planning.
After enforcing controller and joint limits, the robot executes the first action, re-observes, updates phase, and replans.

\section{Experiments}\label{sec:experiments}

\subsection{Experimental Setup}\label{sec:setup}
\noindent\textbf{Simulation Benchmark.}
We evaluate on 30 LIBERO manipulation tasks~\cite{libero} paired with EgoDex egocentric clips~\cite{egodex}, grouped into three difficulty tiers: short-horizon transport (T1), precise placement (T2), and long-horizon rearrangement (T3). Training uses 1,119 paired but unsynchronized human--robot segments; evaluation uses 120 held-out human videos whose source sessions are disjoint from training. Every method runs 3 training seeds and 12 resets per video under frozen success predicates; five ManiSkill task families~\cite{maniskill2} serve as a cross-benchmark testbed without any per-benchmark tuning.

\begin{figure}[!t]
\centering
\subfloat[]{\includegraphics[height=0.2733\textwidth]{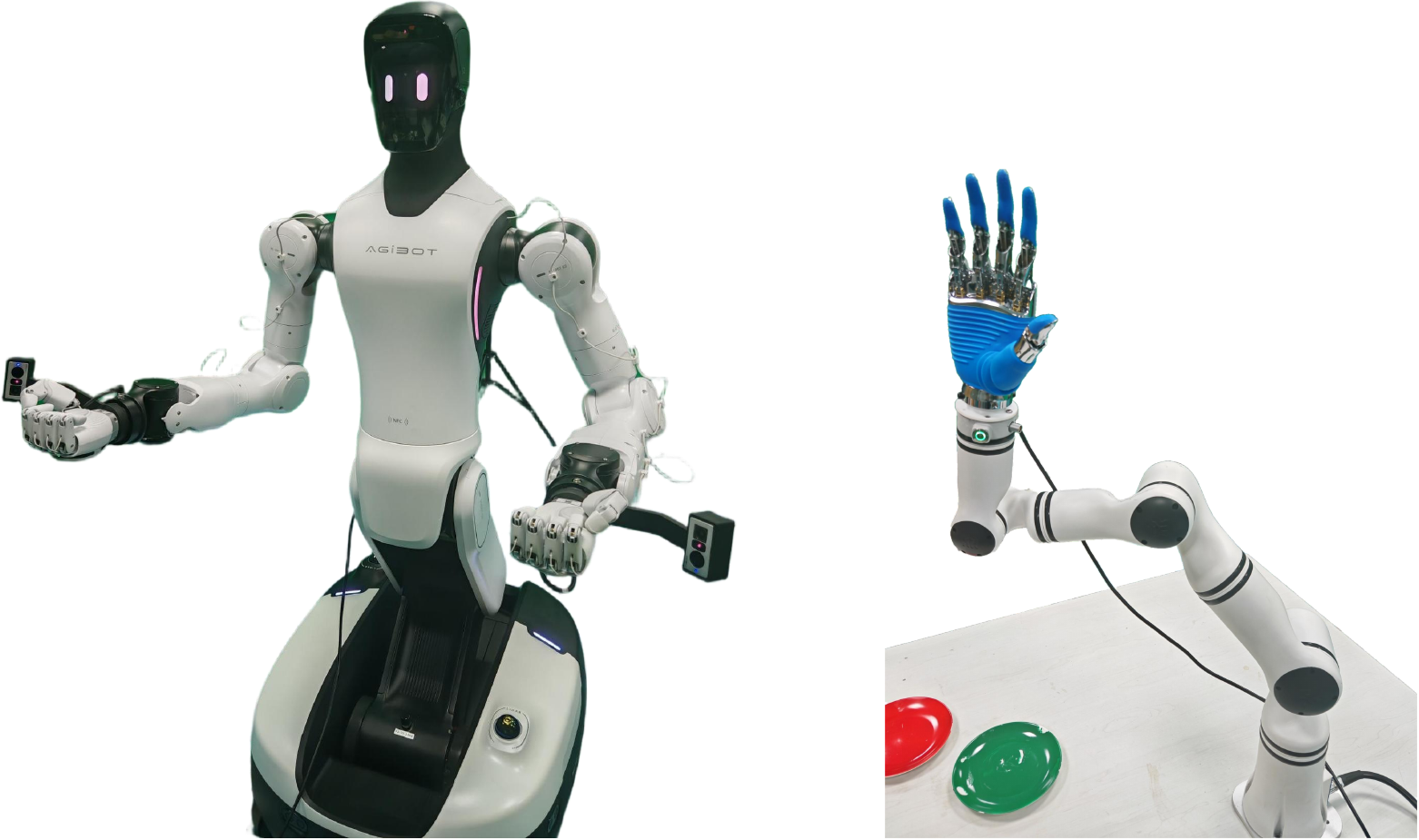}}\hfill
\subfloat[]{\includegraphics[height=0.2733\textwidth]{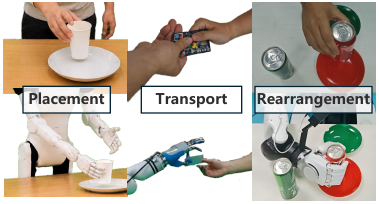}}
\caption{\textbf{Physical-robot zero-shot transfer.}
(a) Physical robot platforms used in the zero-shot transfer experiments. Left: Platform A, a humanoid robot whose 7-DoF arm ends in a dexterous hand. Right: Platform B, a table-mounted 6-DoF manipulator equipped with a dexterous hand. The frozen simulator-selected \cowm{} configuration is deployed on both platforms without per-platform tuning; test human videos are captured in situ and never paired with robot trajectories.
(b) Qualitative zero-shot transfer on the physical platforms. Each column pairs the unseen human demonstration (top) with the corresponding robot execution (bottom) on three representative tabletop tasks, placing a cup onto a plate (\emph{Placement}), handing over a card (\emph{Transport}), and rearranging drinks into the color-matched plates (\emph{Rearrangement}).}
\label{fig:physical}
\end{figure}

\noindent\textbf{Physical-Robot Benchmark.}
The frozen simulator-selected configuration is deployed on a humanoid with a 7-DoF arm (Platform A) and a table-mounted 6-DoF arm (Platform B), each equipped with a dexterous hand (Figure~\ref{fig:physical}(a)); the two platforms differ in kinematic structure and mounting, and each runs the same 5 tabletop tasks with 15 trials per task. Test videos are captured in situ and never paired with robot trajectories.

\noindent\textbf{Baselines.}
\textit{\textbf{Direct retargeting}}: maps retargeted hand poses to end-effector commands without learning;
\textit{\textbf{Phantom}}~\cite{phantom}: edits human videos into robot-compatible observation--action pairs without robot trajectories;
\textit{\textbf{Demo-conditioned BC}}~\cite{bcz}, \textit{\textbf{ACT}}~\cite{act}, and \textit{\textbf{Diffusion Policy}}~\cite{diffusionpolicy}: imitation policies conditioned on the same held-out human video;
\textit{\textbf{MimicPlay}}~\cite{mimicplay} and \textit{\textbf{Track2Act}}~\cite{track2act}: hierarchical and track-based human-to-robot transfer;
\textit{\textbf{LAD}}~\cite{lad}: a diffusion policy in a contrastively aligned cross-embodiment latent action space;
\textit{\textbf{UniSkill}}~\cite{uniskill}: a policy conditioned on embodiment-agnostic skill representations from the human video;
\textit{\textbf{Goal-only MPC}}: our planner driven by the terminal goal alone;
\textit{\textbf{Matched world-model MPC}}: shares the backbone and planner of \cowm but is trained without co-construction.
All learned baselines receive identical human observations, training splits, robot state, and tuning budgets.

\noindent\textbf{Evaluation Metrics.}
\textbf{\textit{Success Rate (SR)}}: the fraction of trials satisfying the frozen task predicate;
\textbf{\textit{Final Error (Err)}}: terminal Euclidean translation error of the manipulated object in meters;
\textbf{\textit{Effect Match (EM)}}: the rate at which the decoded human action reproduces the target object effect from an identical simulator state;
\textbf{\textit{Rejection Rate (RR)}}: the rate at which nearby different-effect actions are correctly separated.
Paired contrasts use a crossed run/source bootstrap with Holm correction (Section~\ref{sec:protocol}).

\noindent\textbf{Implementation Details.} All models are trained on a single server with 8$\times$NVIDIA A800 (80\,GB) GPUs, dual Intel Xeon Platinum CPUs, and 1\,TB RAM running Ubuntu 20.04. The implementation uses Python 3.10, PyTorch 2.1 with CUDA 12.1, and standard scientific libraries (NumPy, SciPy); simulation uses ManiSkill and LIBERO.

\subsection{Evaluation Protocol}\label{sec:protocol}
The following estimands evaluate \cowm{} at shared-space, planner, and task levels. Exact-snapshot simulator interventions and physical-robot trials remain independent evaluation branches; evidence from one branch is not substituted for evidence required in the other. The exact-snapshot intervention and resampling estimands below apply to the controlled LIBERO branch.

\noindent\textbf{Identical-state interventions.} Simulator intervention uses a frozen manifest of tuples $(d,x,i,j,b_=,b_-,s)$: demonstration $d$, reproducible robot snapshot $x$, independently aligned human/robot indices $(i,j)$, equal- and different-effect robot controls $(b_=,b_-)$, and simulator seed $s$. From the encoded snapshot $(\sa_x^r,\pa_x^r)$, the four decoded commands are
\begin{equation}
a^{\rm tar}=A_r(\ua_j^r,\sa_x^r,\pa_x^r),\quad
a^{\rm match}=A_r(\ua_i^h,\sa_x^r,\pa_x^r),\quad
a^{=}=A_r(\ua_{b_=}^r,\sa_x^r,\pa_x^r),\quad
a^{-}=A_r(\ua_{b_-}^r,\sa_x^r,\pa_x^r).
\end{equation}
Each command is executed separately from the exact snapshot with seed $s$ for a validation-frozen duration. Let $\Delta\ra^{\rm tar}$ be the measured object-relation change under $a^{\rm tar}$ and $\Delta\ra^c$ that under candidate $c\in\{\rm match,=,-\}$. Equal/different controls are independently effect-labeled at that snapshot; they do not use the learned SoftDTW path.

For $c\in\{\rm match,=\}$, the match indicator is
\begin{equation}
M_{\rm eff}^{c}=\mathbb I[a^{\rm tar}\text{ and }a^c\text{ feasible}]
\times\mathbb I[\|W_{\rm eff}^{1/2}(\Delta\ra^{c}
-\Delta\ra^{\rm tar})\|_2\leq\epsilon_{\rm eff}].
\end{equation}
For the different-effect control, successful rejection is
\begin{equation}
R_{\rm diff}=\mathbb I[a^{\rm tar}\text{ and }a^-\text{ feasible}]
\times\mathbb I[\|W_{\rm eff}^{1/2}(\Delta\ra^{-}
-\Delta\ra^{\rm tar})\|_2>\epsilon_{\rm eff}].
\end{equation}
$W_{\rm eff}$, action duration, and $\epsilon_{\rm eff}$ are task-specific and fixed on validation data. Infeasible, unmeasurable, tracker-failed, timed-out, or persistently missing candidates receive indicator zero. Rates first average actions, states, and seeds within demonstration and then demonstrations equally.

Let $P_{ab}^{\rm match}$, $P_{ab}^{\rm equal}$, and $P_{ab}^{\rm rej}$ be the demonstration-weighted means of $M_{\rm eff}^{\rm match}$, $M_{\rm eff}^{=}$, and $R_{\rm diff}$ when $a$ toggles $\Lcal_{\rm swap}$ and $b$ toggles $\Lcal_{\rm sens}$. The prespecified mechanism effects are
\begin{equation}
\Delta_{\rm swap}=P_{11}^{\rm match}-P_{01}^{\rm match},
\qquad
\Delta_{\rm sens}=P_{11}^{\rm rej}-P_{10}^{\rm rej}.
\end{equation}
The mechanism claim requires positive $\Delta_{\rm swap}$ and $\Delta_{\rm sens}$ with Holm-adjusted $p<0.05$. For \cowm, one-sided 95\% lower bounds for $P_{11}^{\rm match}$, $P_{11}^{\rm equal}$, and $P_{11}^{\rm rej}$ must exceed validation-frozen $\kappa_{\rm match}$, $\kappa_{\rm equal}$, and $\kappa_{\rm rej}$, respectively. Failure of any criterion withholds the corresponding substitutability or sensitivity claim.

\noindent\textbf{Task success and failure handling.} Let $\bar S_{d,m}$ be binary task success for method $m$, averaged over every preregistered training run and reset for demonstration $d$. The primary contrast is
\begin{equation}
\widehat\Delta_{\rm succ}(m)=\frac{1}{D}\sum_{d=1}^D
(\bar S_{d,\rm ours}-\bar S_{d,m}).
\end{equation}
Here ``ours'' indexes \cowm{} without changing the registered estimand.
A complete run--demonstration--reset and intervention manifest is generated before test access. Training failure, crash, timeout, unsupported task, or output missing beyond a frozen deadline is assigned success zero. A simulator-infrastructure failure before either method acts may be rerun once with the identical seed; persistent failure assigns zero to success, $M_{\rm eff}^{c}$, and $R_{\rm diff}$ and the frozen worst-case value to error metrics. Tracker failure or an unmeasurable intervention is handled identically. No cell is deleted.

\noindent\textbf{Shared-space and planner diagnostics.} On a frozen set $\mathcal A$ of independently aligned tuples $(n,i,j)$, let $t_h=i$ and $t_r=j$. Per-state swap discrepancy is
\begin{equation}
g_{n,ij}^e=\|W_{{\rm eff},n}^{1/2}\operatorname{phys}
(\delta(\sa_{n,t_e}^e,\ua_{n,i}^h)
-\delta(\sa_{n,t_e}^e,\ua_{n,j}^r))\|_2.
\end{equation}
With $Z_{\mathcal A}=2\sum_{(n,i,j)\in\mathcal A}
\widetilde\Gamma_{n,ij}$, the weighted gap is
\begin{equation}
G_{\rm swap}=\frac{1}{Z_{\mathcal A}}
\sum_{(n,i,j)\in\mathcal A}\widetilde\Gamma_{n,ij}
\sum_e g_{n,ij}^e.
\end{equation}
For tuples $\mathcal A^-$ with valid nearby-state negatives, define
\begin{gather}
d_{n,ij,\pm}^{e,W}=\|W_{{\rm eff},n}^{1/2}\operatorname{phys}(
Q_s^{\rm obj}(D_s(\sa_{n,t_e}^e,\ua_{n,ij,\pm}^e))
-\ra_{n,t_e+1,\rm obj}^e)\|_2^2.\\*
G_{\rm neg}=\frac{1}{Z_{\mathcal A^-}}
\sum_{(n,i,j)\in\mathcal A^-}\widetilde\Gamma_{n,ij}
\sum_e(d_{n,ij,-}^{e,W}-d_{n,ij,+}^{e,W}),
\end{gather}
where $Z_{\mathcal A^-}=2\sum_{(n,i,j)\in\mathcal A^-}\widetilde\Gamma_{n,ij}$. $\operatorname{phys}$ reverses training normalization, and task-specific $W_{{\rm eff},n}$ converts translation, rotation, and scalar errors to frozen tolerance units. Lower $G_{\rm swap}$ and higher $G_{\rm neg}$ are better; their intervention counterparts use $M_{\rm eff}^{\rm match}$ and $R_{\rm diff}$.

On the independent phase-annotation subset $\mathcal D_{\rm ph}$, normalized phase error averages time within demonstration and then demonstrations:
\begin{equation}
E_{\rm ph}=\frac{1}{|\mathcal D_{\rm ph}|}
\sum_{d\in\mathcal D_{\rm ph}}\frac{1}{T_d^r}
\sum_t\left|\frac{\hat\tau_{d,t}-\tau_{d,t}^*}{T_d^h}\right|.
\end{equation}
The weighted terminal goal error is $\|W_{{\rm goal},d}^{1/2}\operatorname{phys}(Q_s^{\rm obj}(\sa_{T_d}^r)-\ra_{g,d})\|_2$, a diagnostic quantity in frozen tolerance units; the Err column of Table~\ref{tab:sim-main} and Table~\ref{tab:real-robot}(a) reports only its terminal Euclidean translation component in meters, and rotation and task-scalar components are reported separately in physical units. A failed or missing episode receives a frozen worst-case error. Mask IoU averages valid task-object masks within demonstration and then demonstrations. An identity failure is a role-ID switch or confidence below $\chi_{\min}$ for more than $K_{\rm lost}$ consecutive frames; its rate is events per observed frame, aggregated by demonstration.

Latency reports median and P95 wall-clock time for perception, phase update, proposal, planning, and total control step over all preregistered steps; timeouts enter at the frozen deadline rather than being dropped.

\noindent\textbf{Attribution and human-conditioning controls.} For the $H\times P\times M$ attribution study, the all-zero cell is undefined and omitted. Risk differences give conditional main effects. Two-way interactions are difference-in-differences within the valid stratum of the third factor; no three-way interaction is reported because the all-zero reference does not exist.

The human-conditioning family holds the trained \cowm, robot reset, object identity, terminal goal, and compute budget fixed. The four controls are:
\begin{itemize}
\item \emph{No sequence}: set $H=P=0$ and $M=1$; CEM uses only the terminal-goal, density, smoothness, and feasibility terms.
\item \emph{Shuffled}: apply a manifest-fixed permutation to valid entries of $U^h$ while leaving the human object trajectory, online phase estimate, terminal goal, and padding mask unchanged.
\item \emph{Reversed}: reverse only the valid entries of $U^h$ under the same fixed context.
\item \emph{Alternative}: replace the action and object-state trajectories together with a held-out demonstration having the same named task and object instance, initial relation within $\epsilon_{\rm init}$, and terminal relation within $\epsilon_{\rm goal}$. Before test access, nearest-state donors are assigned one-to-one from a disjoint human-source pool; no donor is reused.
\end{itemize}
All four contrasts use the same frozen cohort of recipients with a valid unique donor. Recipients without a donor remain in the primary intention-to-evaluate analysis but are outside this explicitly scoped conditioning claim; no post-access eligibility change is allowed. The same demonstration-weighted success contrast and crossed bootstrap compare ordered input with each control. Robot-only training is not a conditioning contrast; it belongs to the separately prespecified training-source family below. Any unmatched or post-hoc robot-only run remains exploratory.

\noindent\textbf{Training-source family.} Human-only, robot-only, and joint-source conditions retain the same architecture, optimizer-step budget, train/validation manifests, embodiment-interface calibration, and robot grounding/decoder budget. Only the source allowed to update the shared CoWM state, action, and dynamics modules changes; calibration of the excluded embodiment stops gradients at those modules. The joint-source condition uses both streams but disables $\Lcal_{\rm swap}$ and $\Lcal_{\rm sens}$, separating source co-construction from the complete CoWM semantic constraints.

For the validation-selected human-start and robot-start physical-effect errors, define each primary contrast as single-source error minus joint-source error. The four primary comparisons are joint-source versus human-only and joint-source versus robot-only at each start domain; positive values favor joint-source training. The co-construction claim requires all four point estimates to be positive with Holm-adjusted $p<0.05$. Complete CoWM versus joint-source CoWM is reported separately as an effect-constraint comparison and cannot rescue a failed training-source family. Failed training or interface calibration remains in the intention-to-evaluate analysis under the frozen failure rule.

\noindent\textbf{Statistical analysis.} The confirmatory bootstrap treats training run and human-source cluster as crossed factors. A human-source cluster is the demonstrator ID with all sessions and videos nested beneath it; if demonstrator IDs are unavailable, the highest traceable session ID is used and this limitation is reported. Each replicate samples run and human-source IDs independently with replacement, includes all registered demonstrations in each sampled cluster, resamples matched reset IDs within each run--demonstration cell, and averages resets, demonstrations, and runs in that order with equal demonstration and run weight. The same sampled IDs are applied to both methods. Two-sided percentile intervals are the unadjusted $2.5$th and $97.5$th percentiles of $B=10{,}000$ paired replicates; a one-sided 95\% lower bound is the $5$th percentile. For estimate $\widehat\Delta$ and bootstrap values $\Delta_b^*$, set $z_b=\Delta_b^*-\widehat\Delta$, $N_-=\sum_b\mathbb I[z_b\leq-|\widehat\Delta|]$, and $N_+=\sum_b\mathbb I[z_b\geq|\widehat\Delta|]$. The two-sided value is
\begin{equation}
p=\min\left\{1,2\min\left(
\frac{N_-+1}{B+1},\frac{N_++1}{B+1}\right)\right\},
\end{equation}
followed by separate Holm adjustment within the external-performance family (three contrasts), mechanism-ablation family (two contrasts), human-conditioning family (four contrasts), and training-source family (four contrasts). Every directional claim additionally requires a positive point estimate. The three \cowm{} mechanism thresholds form an intersection rule using their one-sided lower bounds; failure of any threshold fails that mechanism claim. Evidence cannot be borrowed across families. Split sizes, independent human-source count, run/reset counts, power, and all task predicates must be frozen before test access.

\subsection{Main Results}\label{sec:main-results}
Table~\ref{tab:sim-main} reports SR, Err, and EM across three LIBERO
difficulty tiers, while Table~\ref{tab:real-robot}(a) reports SR and Err on the
two physical platforms. EM is restricted to simulation,
where identical-state interventions are reproducible.

\begin{table}[!t]
\centering
\caption{\textbf{Comparison of \cowm with baselines} on 30 LIBERO
    tasks across three difficulty tiers
    (120 held-out human videos, 3 seeds $\times$ 12 resets),
    reporting success rate (SR, \%), final object translation error
    (Err, m), and effect match (EM, \%) measured by identical-state
    simulator interventions. \textbf{Bold} and \underline{underline}
    indicate the best and second-best performance. See Section~\ref{sec:main-results}.}
\label{tab:sim-main}
\footnotesize
\renewcommand{\arraystretch}{1.04}
\setlength{\tabcolsep}{1.15mm}
\adjustbox{max width=\textwidth}{%
\begin{tabular}{c||ccc|ccc|ccc|ccc}
  \Xhline{1.2pt}
  \rowcolor[HTML]{D4D9E9}
  & \multicolumn{3}{c|}{\textbf{T1: Transport}}
  & \multicolumn{3}{c|}{\textbf{T2: Placement}}
  & \multicolumn{3}{c|}{\textbf{T3: Rearrangement}}
  & \multicolumn{3}{c}{\textbf{Overall}} \\
  \rowcolor[HTML]{D4D9E9}
  \multirow{-2}{*}{\textbf{Methods}}
  & SR($\uparrow$) & Err($\downarrow$) & EM($\uparrow$)
  & SR($\uparrow$) & Err($\downarrow$) & EM($\uparrow$)
  & SR($\uparrow$) & Err($\downarrow$) & EM($\uparrow$)
  & SR($\uparrow$) & Err($\downarrow$) & EM($\uparrow$) \\
  \hline
  \hline

  \multicolumn{13}{l}{\textit{Without Robot Learning}} \\
  \hdashline
  \rowcolor[HTML]{F1F3F9}
  Direct retargeting
  & 15.4 & 0.182 & 22.1
  & 6.8 & 0.221 & 15.4
  & 2.7 & 0.254 & 9.8
  & 8.3 & 0.213 & 15.8 \\

  Phantom~\cite{phantom}
  & 34.8 & 0.125 & 38.2
  & 21.6 & 0.147 & 30.5
  & 12.3 & 0.172 & 22.7
  & 22.9 & 0.148 & 30.5 \\
  \hline

  \multicolumn{13}{l}{\textit{With Robot Learning}} \\
  \hdashline
  \rowcolor[HTML]{F1F3F9}
  Demo-conditioned BC~\cite{bcz}
  & 31.5 & 0.131 & 34.6
  & 19.2 & 0.152 & 27.2
  & 11.0 & 0.176 & 19.5
  & 20.6 & 0.147 & 27.1 \\

  Demo-conditioned ACT~\cite{act}
  & 40.3 & 0.108 & 41.8
  & 26.5 & 0.127 & 33.5
  & 16.7 & 0.148 & 25.2
  & 27.8 & 0.121 & 33.5 \\

  \rowcolor[HTML]{F1F3F9}
  Diffusion Policy~\cite{diffusionpolicy}
  & 45.1 & 0.097 & 46.3
  & 30.8 & 0.114 & 38.1
  & 19.8 & 0.139 & 29.6
  & 31.9 & 0.109 & 38.0 \\

  MimicPlay~\cite{mimicplay}
  & 49.6 & 0.091 & 50.7
  & 34.7 & 0.107 & 42.4
  & 22.4 & 0.128 & 33.8
  & 35.6 & 0.102 & 42.3 \\

  \rowcolor[HTML]{F1F3F9}
  Track2Act~\cite{track2act}
  & 52.8 & 0.084 & 54.2
  & 37.9 & 0.099 & 46.0
  & 24.1 & 0.121 & 36.9
  & 38.3 & 0.095 & 45.7 \\

  LAD~\cite{lad}
  & 54.7 & 0.081 & 56.1
  & 39.8 & 0.096 & 47.7
  & 26.3 & 0.118 & 38.4
  & 40.3 & 0.092 & 47.4 \\

  \rowcolor[HTML]{F1F3F9}
  UniSkill~\cite{uniskill}
  & 56.4 & 0.078 & 58.3
  & 41.5 & 0.093 & 49.6
  & 27.9 & 0.114 & 40.8
  & 41.9 & 0.089 & 49.6 \\
  \hline

  \multicolumn{13}{l}{\textit{Ours Design}} \\
  \hdashline
  \rowcolor[HTML]{F1F3F9}
  Goal-only MPC
  & 36.2 & 0.118 & \underline{84.2}
  & 23.4 & 0.139 & \underline{79.1}
  & 14.6 & 0.163 & \underline{74.9}
  & 24.7 & 0.133 & \underline{79.4} \\

  Matched world-model MPC
  & \underline{58.9} & \underline{0.069} & 44.6
  & \underline{45.2} & \underline{0.081} & 38.9
  & \underline{33.4} & \underline{0.102} & 33.2
  & \underline{45.8} & \underline{0.078} & 38.9 \\

  \rowcolor[HTML]{FDF6DA}
  \textbf{\cowm~(Ours)}
  & \textbf{76.5} & \textbf{0.036} & \textbf{86.1}
  & \textbf{63.0} & \textbf{0.043} & \textbf{81.0}
  & \textbf{53.2} & \textbf{0.055} & \textbf{77.2}
  & \textbf{64.2} & \textbf{0.041} & \textbf{81.4} \\
  \Xhline{1.2pt}
\end{tabular}}
\end{table}

\begin{table}[!t]
\centering
\caption{\textbf{Zero-shot transfer on the two physical platforms.}
(a) Comparison with baselines (5 tasks $\times$ 15 trials each); test videos receive no robot trajectory, teleoperation, or scripted control.
(b) Per-task success counts of \cowm{} on the two physical platforms (15 trials per task).
\textbf{Bold} and \underline{underline} indicate the best and second-best performance. See Section~\ref{sec:main-results}.}
\label{tab:real-robot}
\footnotesize
\renewcommand{\arraystretch}{1.15}
\subfloat[]{\setlength{\tabcolsep}{9.06pt}%
\begin{tabular}{c||cc|cc}
  \Xhline{1.2pt}
  \rowcolor[HTML]{D4D9E9}
  & \multicolumn{2}{c|}{\textbf{Robot A (7-DoF)}} & \multicolumn{2}{c}{\textbf{Robot B (6-DoF)}} \\
  \rowcolor[HTML]{D4D9E9}
  \multirow{-2}{*}{\textbf{Methods}} & SR($\uparrow$) & Err($\downarrow$) & SR($\uparrow$) & Err($\downarrow$) \\
  \hline
  \hline
  \rowcolor[HTML]{F1F3F9}
  Direct retargeting & 9.3 & 0.201 & 6.7 & 0.224 \\
  Diffusion Policy & 26.7 & 0.118 & 22.7 & 0.131 \\
  \rowcolor[HTML]{F1F3F9}
  Track2Act & 32.0 & 0.104 & 28.0 & 0.117 \\
  UniSkill & 34.7 & 0.098 & 30.7 & 0.109 \\
  \rowcolor[HTML]{F1F3F9}
  Matched WM MPC & \underline{37.3} & \underline{0.089} & \underline{33.3} & \underline{0.098} \\
  \rowcolor[HTML]{FDF6DA}
  \textbf{\cowm~(Ours)} & \textbf{66.7} & \textbf{0.046} & \textbf{61.3} & \textbf{0.053} \\
  \Xhline{1.2pt}
\end{tabular}}\hfill
\subfloat[]{\setlength{\tabcolsep}{9.06pt}%
\begin{tabular}{c||cc}
  \Xhline{1.2pt}
  \rowcolor[HTML]{D4D9E9}
  & & \\
  \rowcolor[HTML]{D4D9E9}
  \multirow{-2}{*}{\textbf{Task}} & \multirow{-2}{*}{\textbf{Platform A}} & \multirow{-2}{*}{\textbf{Platform B}} \\
  \hline
  \hline
  \rowcolor[HTML]{F1F3F9}
  Placement & 12/15 & 11/15 \\
  Transport & 12/15 & 11/15 \\
  \rowcolor[HTML]{F1F3F9}
  Rearrangement & 9/15 & 8/15 \\
  Stacking & 10/15 & 9/15 \\
  \rowcolor[HTML]{F1F3F9}
  Insertion & 7/15 & 7/15 \\
  \rowcolor[HTML]{FDF6DA}
  Total & 50/75 (66.7\%) & 46/75 (61.3\%) \\
  \Xhline{1.2pt}
\end{tabular}}
\end{table}

\begin{figure}[!t]
\centering
\includegraphics[width=0.94\textwidth]{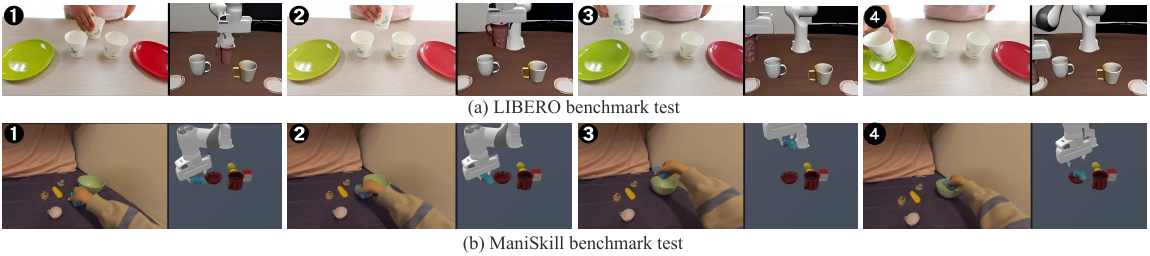}
\caption{\textbf{Qualitative zero-shot transfer in simulation.} Each numbered pair shows the unseen human demonstration (left) and the corresponding robot execution (right) at matched task phases on (a) the LIBERO benchmark and (b) the ManiSkill benchmark. Given one unseen human video, \cowm tracks the demonstrated object-relation evolution in the shared latent space and reproduces it with its own embodiment in both simulators (see the supplementary video for complete rollouts).}
\label{fig:qualitative}
\end{figure}

\noindent\textbf{Effectiveness of \cowm.}
Across tiers, \cowm achieves the best SR and Err, reaching 64.2\% overall success versus 45.8\% for the best baseline while halving the final object error. Recent skill- and action-latent methods narrow the gap among baselines (UniSkill 41.9\%, LAD 40.3\%) yet remain 22 points behind, since their latents are aligned by future-frame prediction or contrastive kinematic retargeting rather than bound to predicted object effects; Phantom transfers visually edited demonstrations without robot learning and thus saturates earlier. The margin widens with task difficulty (T3: +19.8 points over matched world-model MPC and +25.3 over UniSkill), where preserving intermediate interaction dynamics matters most; the gap to the matched world-model MPC, which shares the backbone and planner, isolates the contribution of co-construction and the effect constraints. Figure~\ref{fig:qualitative} visualizes phase-matched rollouts on LIBERO and ManiSkill, where \cowm reproduces the demonstrated object-relation evolution from an unseen video.

\begin{figure}[!t]
\centering
\subfloat[]{\includegraphics[height=0.2219\textwidth]{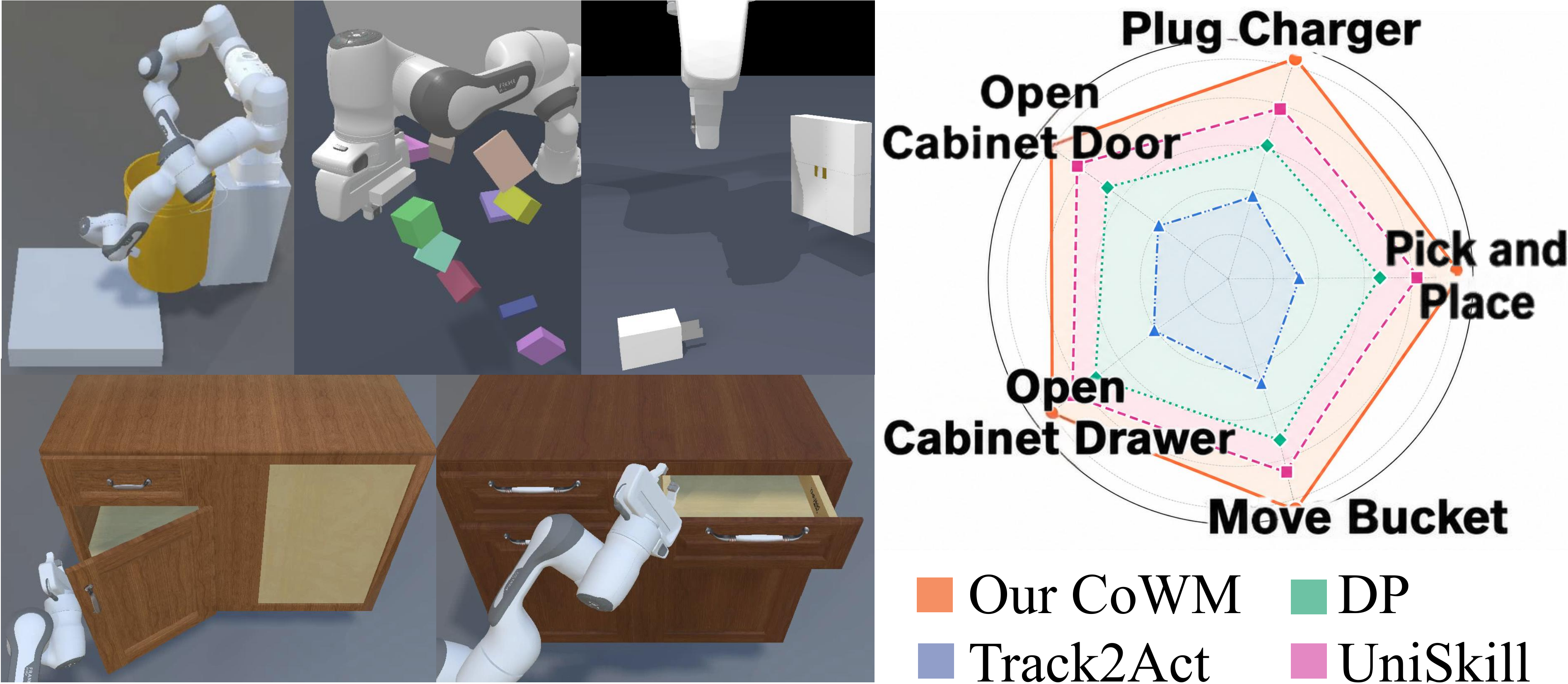}}\hfill
\subfloat[]{\includegraphics[height=0.2219\textwidth]{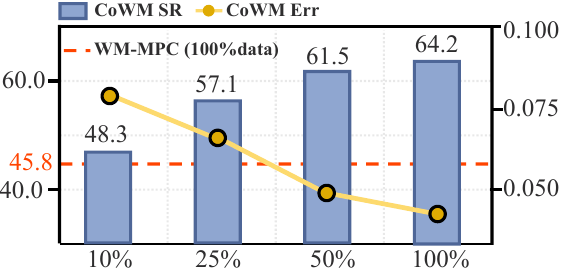}}
\caption{\textbf{Cross-benchmark transfer and robot-data efficiency.}
(a) Results on the ManiSkill benchmark. Success rates on five ManiSkill task families under the LIBERO-selected configuration without per-benchmark tuning.
(b) Robot-data efficiency with a fixed human stream. Bars give SR with 10--100\% of the robot data, the curve the final object error, and the dashed line the matched world-model MPC on full data.}
\label{fig:maniskill}
\end{figure}

\noindent\textbf{Cross-Benchmark Transfer on ManiSkill.}
To test whether the co-constructed space is tied to one simulator, we transfer the LIBERO-selected configuration to the ManiSkill benchmark without per-benchmark tuning (Figure~\ref{fig:maniskill}(a)). \cowm attains 81.24\% SR on tabletop pick-and-place and degrades gracefully on the harder families (27.78\% OpenCabinetDrawer, 21.34\% CabinetDoor, 11.58\% MoveBucket, 10.32\% PlugCharger), where articulated, whole-body, and millimeter-precision interaction exceeds the tabletop training regime, yet it leads UniSkill, Track2Act, and Diffusion Policy on every family, so effect-grounded latents carry across benchmarks and the residual gap tracks interaction complexity rather than embodiment.

\begin{figure}[!t]
\centering
\includegraphics[width=0.78\textwidth]{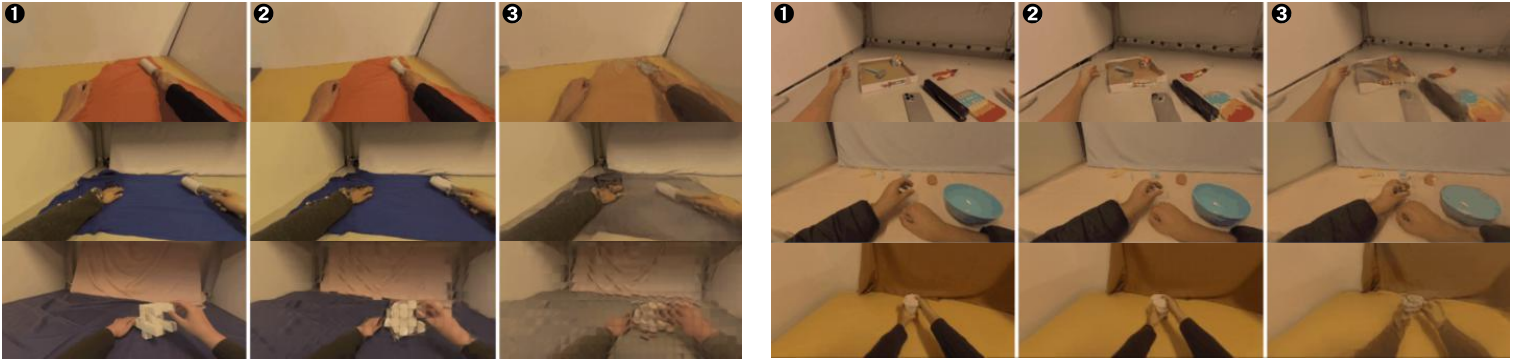}
\caption{\textbf{Visualization of world-model prediction in the DINOv2 feature space.} Each of the two panels shows three held-out sequences (rows). Column \ding{182}: the ground-truth observation (GT). Column \ding{183}: the reconstruction decoded from the latent encoded directly from the observed goal frame (Goal-latent recon), which bounds what the decoder can recover. Column \ding{184}: the reconstruction decoded from the latent rolled out by the shared dynamics $D_s$ from an earlier state and the intervening latent actions (Pred-latent recon). The predicted latents preserve the object configuration and contact structure of the ground truth at a fidelity close to the goal-latent bound, so the rollout errors of the shared dynamics remain small in the task-relevant content rather than only in feature norm.}
\label{fig:dino-recon}
\end{figure}

\noindent\textbf{Effect-Level Transfer.}
EM dissociates transferred action semantics from execution. Figure~\ref{fig:dino-recon} visualizes the shared dynamics by decoding both encoded and rolled-out latents back to image space, separating decoder capacity (middle column) from prediction quality (right column). \cowm reaches 81.4\% overall EM against 49.6\% for the best external baseline (UniSkill), and the two variants fail differently: Goal-only MPC shares the co-constructed model and retains near-\cowm EM (79.4\%) despite its low SR, so its failures stem from discarding intermediate trajectory guidance, whereas the matched world-model MPC attains the second-best SR with only 38.9\% EM, so planning capacity cannot compensate for unbound action semantics. Baseline EM tracks the SR ordering and degrades with tier difficulty, consistent with effect binding driving transfer.

\begin{figure}[!t]
\centering
\includegraphics[width=0.86\textwidth]{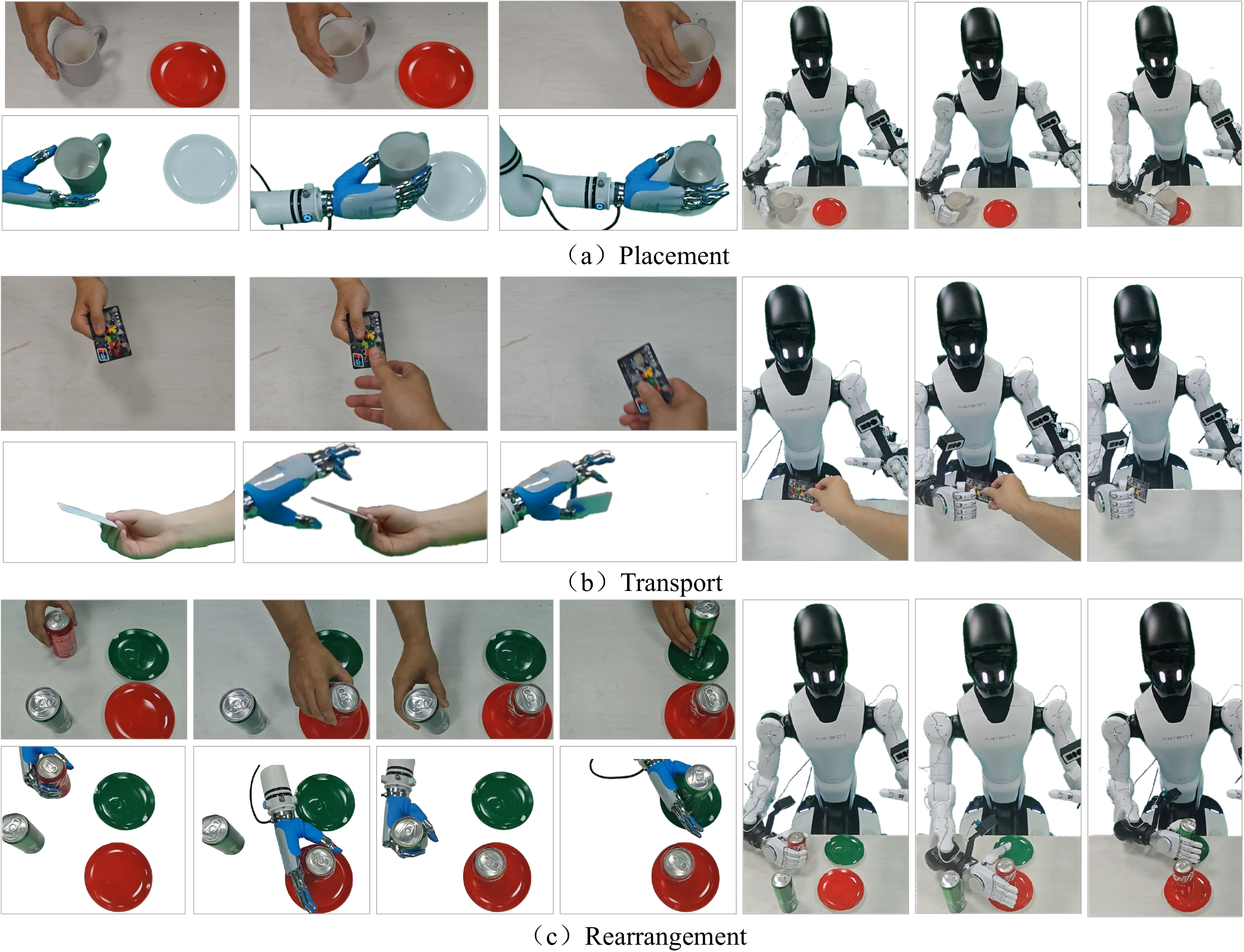}
\caption{\textbf{Real-robot rollouts on three representative tasks.} (a) \emph{Placement}: placing a cup onto a plate; (b) \emph{Transport}: handing over a card; (c) \emph{Rearrangement}: rearranging drinks into the color-matched plates. In each panel, the left block shows the human demonstration (top row) and the task executed by Platform B (bottom row), and the right block shows the task executed by Platform A.}
\label{fig:real-exp}
\end{figure}

\noindent\textbf{Physical-Robot Transfer.}
On both platforms \cowm roughly doubles the strongest baseline (66.7\% and 61.3\% SR), and the method ranking is consistent with simulation, so the co-constructed latent space rather than platform-specific tuning drives the transfer. Figure~\ref{fig:physical}(b) and Figure~\ref{fig:real-exp} show representative rollouts on both platforms, where the robots reproduce the cup placement, card hand-over, and color-matched drink rearrangement after watching the human once. Table~\ref{tab:real-robot}(b) lists the per-task success counts of \cowm{} behind these aggregate success rates (Platform A 50/75, Platform B 46/75), and Table~\ref{tab:diagnostics}(a) categorizes the 54 failed trials. Each failed trial receives exactly one label under a frozen first-failure precedence rule (perception $\rightarrow$ phase $\rightarrow$ planning $\rightarrow$ grasp $\rightarrow$ collision $\rightarrow$ timeout), so the categories are mutually exclusive and the shares sum to one.

\begin{table}[!t]
\centering
\caption{\textbf{Failure taxonomy and component ablation.}
(a) Failure taxonomy over the 54 failed physical-robot trials; labels are mutually exclusive under the frozen first-failure precedence rule.
(b) Ablation on execution and representation components; training-side components are removed by retraining, execution-side ones are disabled at test time.
\textbf{Bold} and \underline{underline} indicate the best and second-best performance. See Sections~\ref{sec:main-results} and~\ref{sec:ablation}.}
\label{tab:diagnostics}
\footnotesize
\renewcommand{\arraystretch}{1.15}
\subfloat[]{\setlength{\tabcolsep}{5.37pt}%
\begin{tabular}{cc||cc|c}
  \Xhline{1.2pt}
  \rowcolor[HTML]{D4D9E9}
  \textbf{Category} & \textbf{Description} & \textbf{Platform A} & \textbf{Platform B} & Share \\
  \hline
  \hline
  \rowcolor[HTML]{F1F3F9}
  Perception & Tracking loss & 4 & 5 & 16.7\% \\
  Phase & Progress mismatch & 3 & 4 & 13.0\% \\
  \rowcolor[HTML]{F1F3F9}
  Planning & No feasible elite & 4 & 4 & 14.8\% \\
  Grasp & Slip or missed contact & 6 & 8 & 25.9\% \\
  \rowcolor[HTML]{F1F3F9}
  Collision & Safety stop & 4 & 5 & 16.7\% \\
  Timeout & -- & 4 & 3 & 13.0\% \\
  \rowcolor[HTML]{FDF6DA}
  \multicolumn{2}{c||}{Total} & 25 & 29 & 100\% \\
  \Xhline{1.2pt}
\end{tabular}}\hfill
\subfloat[]{\setlength{\tabcolsep}{5.37pt}%
\begin{tabular}{c||cc}
  \Xhline{1.2pt}
  \rowcolor[HTML]{D4D9E9}
  \textbf{Variants} & SR($\uparrow$) & Err($\downarrow$) \\
  \hline
  \hline
  \rowcolor[HTML]{F1F3F9}
  w/o embodiment-private context & 49.6 & 0.078 \\
  w/o action-support density & 51.4 & 0.072 \\
  \rowcolor[HTML]{F1F3F9}
  w/o phase tracking & 52.8 & 0.069 \\
  w/o residual proposal & 55.3 & 0.063 \\
  \rowcolor[HTML]{F1F3F9}
  w/o seed initialization & 58.9 & 0.052 \\
  w/o MK-MMD & \underline{61.7} & \underline{0.047} \\
  \rowcolor[HTML]{FDF6DA}
  \textbf{\cowm~(Ours)} & \textbf{64.2} & \textbf{0.041} \\
  \Xhline{1.2pt}
\end{tabular}}
\end{table}

\subsection{Ablation Study}\label{sec:ablation}
\noindent\textbf{Effect Constraints.}
We toggle $\Lcal_{\rm swap}$ and $\Lcal_{\rm sens}$ in a prespecified $2\times2$ design and evaluate identical-state robot interventions (Table~\ref{tab:ablation}(a)). Adding $\Lcal_{\rm swap}$ to the sensitivity-only variant raises EM from 46.8\% to 81.4\%, adding $\Lcal_{\rm sens}$ to the swap-only variant raises RR from 15.3\% to 74.6\%, and only their combination attains both, adding at least 12.5 SR points over either alone. Both prespecified effects are positive at Holm-adjusted $p<0.05$: swap binds matched effects across embodiments, and sensitivity rejects the action-agnostic shortcut. On the frozen test cohort with $B=10{,}000$ crossed run--source replicates, the observed effects are $\Delta_{\rm swap}=+0.346$ ($0.814-0.468$; Holm-adjusted $p=0.004$, one-sided 95\% lower bound $+0.281$) and $\Delta_{\rm sens}=+0.593$ ($0.746-0.153$; Holm-adjusted $p=0.002$, one-sided 95\% lower bound $+0.512$). The validation-frozen thresholds are $\kappa_{\rm match}=0.70$, $\kappa_{\rm equal}=0.60$, and $\kappa_{\rm rej}=0.60$, and the one-sided 95\% lower bounds for $P_{11}^{\rm match}$, $P_{11}^{\rm equal}$, and $P_{11}^{\rm rej}$ are $0.769$, $0.716$, and $0.688$, so all three criteria are met.

\begin{table}[!t]
\centering
\caption{\textbf{Effect-constraint ablation and timing robustness.}
(a) Ablation on the effect constraints via identical-state interventions (rates in \%), highlighting their complementary, non-redundant roles.
(b) Robustness to timing perturbations of the human demonstration, highlighting the contribution of phase-aware planning under execution-rate mismatch.
\textbf{Bold} and \underline{underline} indicate the best and second-best performance. See Sections~\ref{sec:ablation} and~\ref{sec:timing}.}
\label{tab:ablation}
\footnotesize
\renewcommand{\arraystretch}{1.15}
\subfloat[]{\setlength{\tabcolsep}{10.18pt}%
\begin{tabular}{cc||ccc}
  \Xhline{1.2pt}
  \rowcolor[HTML]{D4D9E9}
  $\Lcal_{\rm swap}$ & $\Lcal_{\rm sens}$ & EM($\uparrow$) & RR($\uparrow$) & SR($\uparrow$) \\
  \hline
  \hline
  \rowcolor[HTML]{F1F3F9}
  \textcolor{red(pigment)}{\xmark} & \textcolor{red(pigment)}{\xmark} & 41.2 & 12.5 & 39.4 \\
  \textcolor{green(pigment)}{\cmark} & \textcolor{red(pigment)}{\xmark} & \underline{78.6} & 15.3 & \underline{51.7} \\
  \rowcolor[HTML]{F1F3F9}
  \textcolor{red(pigment)}{\xmark} & \textcolor{green(pigment)}{\cmark} & 46.8 & \underline{71.2} & 47.5 \\
  \rowcolor[HTML]{FDF6DA}
  \textcolor{green(pigment)}{\cmark} & \textcolor{green(pigment)}{\cmark} & \textbf{81.4} & \textbf{74.6} & \textbf{64.2} \\
  \Xhline{1.2pt}
\end{tabular}}\hfill
\subfloat[]{\setlength{\tabcolsep}{10.18pt}%
\begin{tabular}{c||ccc}
  \Xhline{1.2pt}
  \rowcolor[HTML]{D4D9E9}
  \textbf{Perturbation} & Phase Err($\downarrow$) & SR($\uparrow$) & SR w/o phase($\uparrow$) \\
  \hline
  \hline
  \rowcolor[HTML]{F1F3F9}
  Nominal & 0.013 & \textbf{64.2} & 52.8 \\
  Speed change & 0.035 & 62.4 & 45.1 \\
  \rowcolor[HTML]{F1F3F9}
  Pause & 0.038 & 60.9 & 40.1 \\
  Delayed contact & 0.027 & 61.8 & 43.6 \\
  \Xhline{1.2pt}
\end{tabular}}
\end{table}

\noindent\textbf{Execution Components.}
Table~\ref{tab:diagnostics}(b) removes each component from the full system. The embodiment-private context contributes most ($-$14.6 SR), as removing it forces reconstruction through the shared state. The action-support density follows ($-$12.8): without it the planner exploits infeasible commands. Phase tracking matters chiefly on long-horizon tasks with speed mismatch ($-$11.4 overall); the residual proposal and seed initialization give smaller consistent gains, and MK-MMD is a weak regularizer.

\subsection{Robot-Data Efficiency}\label{sec:data-efficiency}
To assess how far human video substitutes for robot trajectories, we fix the human stream and train \cowm with 10--100\% of the robot data (Figure~\ref{fig:maniskill}(b)). With 25\% of the robot data, \cowm attains 57.1\% SR, exceeding the matched world-model MPC trained on the full set (45.8\%), with marginal returns beyond 50\%; co-construction thus converts cheap human observations into robot competence.

\subsection{Robustness to Timing Perturbations}\label{sec:timing}
We perturb held-out human videos with speed changes, pauses, and delayed contact, and compare the complete \cowm{} with a variant that replaces phase-aware planning by elapsed-time indexing (Table~\ref{tab:ablation}(b)). Phase error stays within 0.04 and SR degrades by at most 3.3 points, whereas the elapsed-time variant loses 12.7 points under pauses; the median planning latency of 11.3~ms per control step supports real-time closed-loop execution.

\section{Conclusion}\label{sec:conclusion}
We introduced the \emph{\cowmlong}, co-constructed from human observations and robot experience, which shares object-relation state, latent actions, and task dynamics across embodiments while retaining private context. Effect-swap and action-sensitivity constraints bind both embodiments' actions through predicted object effects, and phase-aware planning grounds the plan in executable control without any test-task robot trajectory, enabling zero-shot skill transfer. Heavy occlusion, absent force sensing, a global action-support density, and dexterous or deformable-object manipulation remain open limitations.


\Supplements{Video demonstrating complete zero-shot rollouts in simulation and on the two physical platforms.}

\bibliographystyle{scis}
\bibliography{references}

\end{document}